\documentclass{article}
\usepackage{spconf,amsmath,graphicx}
\usepackage{color}
\usepackage{amsfonts}
\usepackage{booktabs}
\usepackage{multirow}
\usepackage[misc]{ifsym}
\usepackage{algorithm}
\usepackage{algpseudocode}
\usepackage{color}
\usepackage{hyperref}

\title{Asymmetric Polynomial Loss For Multi-Label Classification}

\name{Yusheng Huang, Jiexing Qi, Xinbing Wang, Zhouhan Lin $^{\textrm{\Letter}}$ \thanks{${\textrm{\Letter}}$ Zhouhan Lin is the corresponding author.}}
\address{Shanghai Jiao Tong University. Shanghai, China}

\begin{document}
\maketitle
\begin{abstract}
Various tasks are reformulated as multi-label classification problems, in which the binary cross-entropy (BCE) loss is frequently utilized for optimizing well-designed models.
However, the vanilla BCE loss cannot be tailored for diverse tasks, resulting in a suboptimal performance for different models.
Besides, the imbalance between redundant negative samples and rare positive samples could degrade the model performance.
In this paper, we propose an effective Asymmetric Polynomial Loss (APL) to mitigate the above issues. 
Specifically, we first perform Taylor expansion on BCE loss. Then we ameliorate the coefficients of polynomial functions. 
We further employ the asymmetric focusing mechanism to decouple the gradient contribution from the negative and positive samples. Moreover, we validate that the polynomial coefficients can recalibrate the asymmetric focusing hyperparameters.
Experiments on relation extraction, text classification, and image classification show that 
our APL loss can consistently improve performance without extra training burden.\footnote{We release our code at \url{https://github.com/LUMIA-Group/APL}.}
\end{abstract}
\section{Introduction}
Many tasks, including text classification \cite{DBLP:conf/sigir/LiuCWY17} and image classification \cite{DBLP:conf/cvpr/LanchantinWOQ21,DBLP:conf/cvpr/0002ZFY019}, can be formulated into multi-label classification problems, and BCE loss is often used as the training objective.
Specifically, the multi-label classification problem is reduced to a series of independent binary classification subproblems, and
in each of them the negative log-likelihood loss is optimized.

However, since recent researches show that BCE loss could be a suboptimal solution for different tasks \cite{DBLP:conf/aaai/ZhaoYR0W022,DBLP:conf/aaai/Zhou0M021,DBLP:conf/iccv/RidnikBZNFPZ21,DBLP:conf/ijcai/ZhouL22,chen2022transhash,chen2022supervised}, we aim to design a simply tuned loss function with a larger design space that can be tailored for various tasks.
Specifically, we focus on two aspects.
Firstly, for each independent binary classification subproblem, the negative log-likelihood loss is suboptimal in terms of the polynomial coefficients of the logarithm's Taylor expansion \cite{DBLP:journals/corr/abs-2204-12511}. 
This result is also consistent with the optimization of each independent label in the multi-label classification problem. 
Secondly, the imbalance between the number of rare positive labels and redundant negative labels is an obstacle to multi-label classification \cite{DBLP:conf/iccv/LinGGHD17}, and consequently, BCE loss is a suboptimal solution for learning the features of positive samples \cite{DBLP:conf/iccv/RidnikBZNFPZ21}. 
Because the massive negative samples contribute more gradient weights than the fewer positve samples, 
neural networks may focus on learning features from negative samples and pay less attention to positive samples. 
This will lead to a low confidence in the prediction of positive samples. 
The models would under-emphasize the gradient of positive samples and get suboptimal solutions, which could degrade the model performance.

In this paper, we propose an Asymmetric Polynomial Loss (APL) to counteract these deficiencies and improve the performance of multi-label classification.
As stated before, the coefficients of the leading polynomials are suboptimal, and we propose to adjust them for different tasks and datasets. 
While a prominent question is the number of classes may be rather large,
it is impracticable to adjust the coefficient for each class separately. Therefore, we propose to collectively tune the coefficients of leading polynomial bases for all classes, which makes the parameter tuning feasible. 

To mitigate the positive-negative imbalance issue, we further exploit the asymmetric focusing mechanism \cite{DBLP:conf/iccv/RidnikBZNFPZ21} proposed in the ASL loss to enhance the gradient contribution from rare positive samples, which aims to improve the feature learning. 
Specifically, we set different focusing parameters to the Taylor expansion of the BCE loss.
Different focusing parameters enable our loss to decouple the gradient contribution, and we could emphasize the positive samples.
To further reduce the interference of easy negative samples, 
we discard them by excluding them from the calculation.
Furthermore, we have validated theoretically and experimentally that the polynomial coefficients of Taylor expansion maintain the ability to recalibrate the parameters of the asymmetric focusing mechanism. 

In summary, our contributions include:

(1) We propose an asymmetric polynomial loss APL from the perspective of Taylor expansion, which could be tailored for various tasks.
(2) We analyze the APL properties and elaborate on parameter correlations between polynomial coefficients and asymmetric focusing.
(3) We conduct experiments on four datasets, and APL achieves superior performance, which includes state-of-the-art results on MS-COCO and NUS-WIDE.

\section{Methodology}
\subsection{Taylor Expansion for BCE}\label{S1}
The BCE loss could be decomposed into $C$ independent binary classification subproblems $L_{\rm BCE} = \sum_{i=0}^{C} (y_iL_++(1-y_i)L_-)/C$,  $y_i \in \{1, 0\}$,  where $L_+ = -\log (p_i)$ is for the positive class and $L_- = -\log (1-p_i)$ is for the negative class. $p_i$ is the prediction probability after the sigmoid function. 

Since we aim to optimize the subproblems of BCE and adjust the polynomial coefficients following \cite{DBLP:journals/corr/abs-2204-12511}, we firstly apply Taylor series expansion \cite{linnainmaa1976taylor} on $L_{\rm BCE}$. 
For positive classes where $y_i=1$, we set the expansion point to be $1$, and for negative classes where $y_i=0$, we set the expansion point to be $0$.
Thus, Taylor series for BCE loss is 
$
L_{\rm t-BCE} = \sum_{i=0}^{C} [ y_i \sum_{m=1}^{\infty} \alpha_{i, m} (1-p_i)^{m} + (1-y_i) \sum_{n=1}^{\infty} \beta_{i, n} p_i^n ]/C,
$
where $\alpha_{i, m} = \frac{1}{m}$, $\beta_{i, n} = \frac{1}{n}$, $m, n \in \mathbb{N}_+$ are polynomial coefficients.
We would amend the leading polynomial coefficients such as $\alpha_{i, 1}, \alpha_{i, 2}, \beta_{i, 1}$ because they contribute a higher proportion of gradients in the later training phase. 

\subsection{Asymmetric Polynomial Loss}\label{S2}
To mitigate the positive-negative imbalance problem, we scale the loss of positive and negative classes to prevent massive negative samples from overwhelming the training process. 
Specifically, to differentiate the positive and negative classes, we propose to leverage the asymmetric focusing mechanism that employs distinct scaling factors $\gamma^{+/-}$ to $L_{+/-}$. In this way, we can adjust the order of starting polynomials of $L_+$ and $L_-$, which enables us to manually change the gradient proportions of the positive and negative classes. 

To further reduce the influence of easy negative classes on model optimization, we discard these negative classes with small prediction probabilities. 
In detail, we prohibit them with predicted probability $p_i$ lower than the threshold $p_{th}$ from participating in the calculation of loss, i.e.,
$
L_- =  \sum_{n=1}^{\infty} \beta_{i, n} {\text{max}(p_i - p_{th}, 0) }^{n+\gamma^-},
$
where $p_{th} > 0$ is a hard threshold. 

Based on the facts that individually tuning parameters is prohibitive and amending the leading polynomial coefficients leads to maximal gain, we propose a simple implementation that collectively adjusts coefficients of leading polynomial bases $(1-p_i)^{1\&2}$ for all positive/negative classes.
Aggregating the above techniques, our proposed Asymmetric Polynomial Loss (APL) is formalized as follows, 

\noindent$
  L_{\rm APL} = \sum_{i=0}^{C} [ y_i \sum_{m=1}^{\infty} \alpha_{i, m} (1-p_i)^{m+\gamma^+} + \\
\indent \indent \ \ \ (1-y_i) \sum_{n=1}^{\infty} \beta_{i, n}  {\text{max}(p_i - p_{th}, 0) }^{n+\gamma^-} ]/C \\
 = \sum_{i=0}^{C} \{  y_i(1-p_i)^{\gamma^+} [ -\log p_i + (\alpha_1-1)(1-p_i)
 + (\alpha_2-\frac{1}{2})(1-p_i)^2 ] + (1-y_i)p_{res}^{\gamma^-}
 [ -\log (1-p_i) + (\beta_1-1)p_{res} ] \}/C, 
$

\noindent where $\alpha_1=\alpha_{i, 1}, \alpha_2=\alpha_{i, 2}, \beta_1=\beta_{i,1}, $ $i=1,2,...,C$ are coefficients, and $p_{res}$ denotes $\text{max}(p_i - p_{th}, 0)$.

\begin{figure}[t]
\centering
\includegraphics[width=0.35\textwidth]{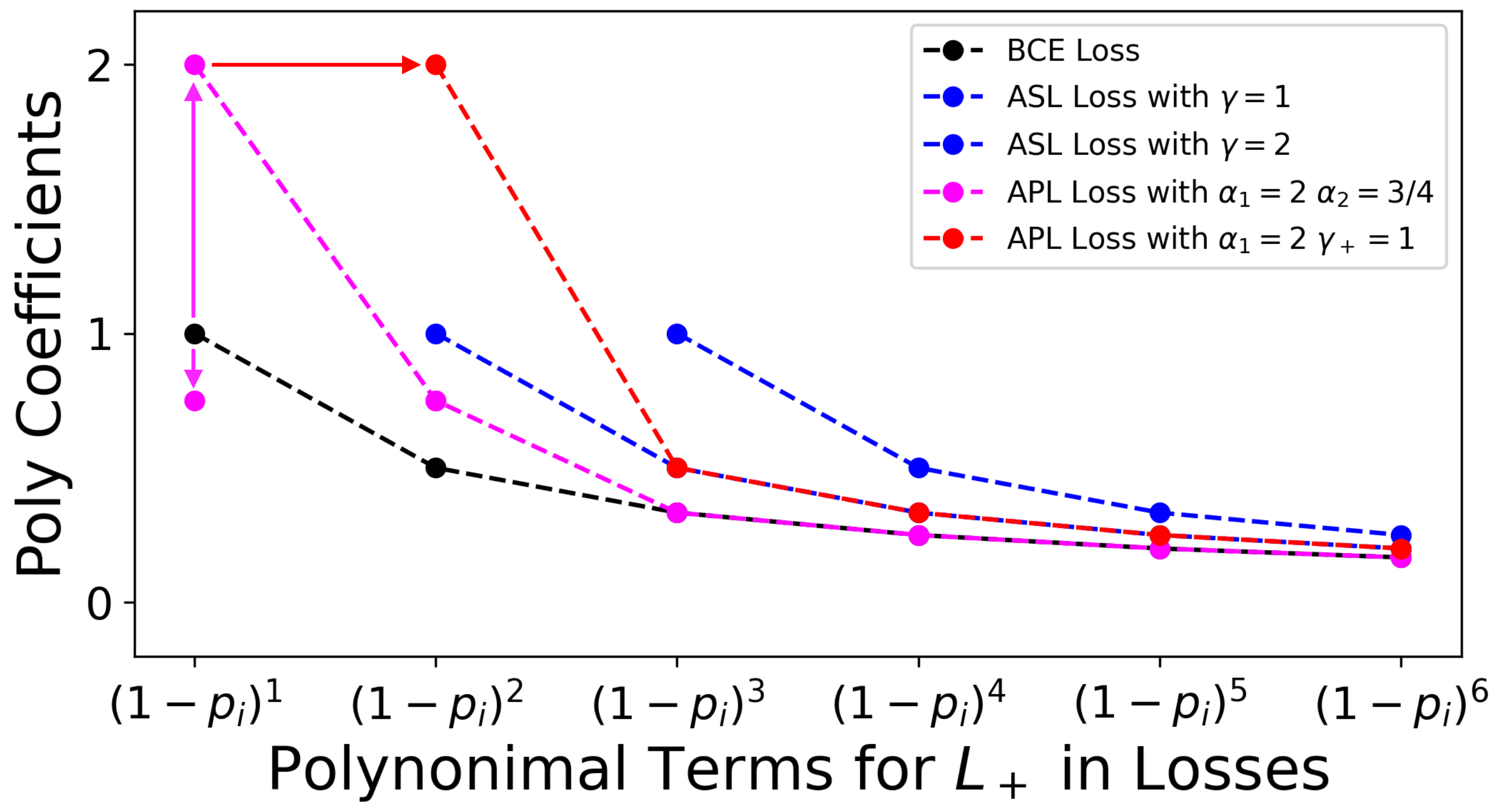}
\caption{Polynomial coefficients of $L_+$ loss for different loss functions in the base of $(1-p_i)^k, k \in \mathbb{N}+$.}
\label{pic_poly}
\end{figure}

Our APL loss maintains good flexibility in the choice of parameters for the polynomial basis. As presented in Figure \ref{pic_poly}, compared with BCE loss with fixed coefficients and ASL loss that can only be shifted horizontally, our APL loss allows for simultaneous vertical and horizontal adjustments, which greatly increases the design space for miscellaneous tasks. So $L_-$ loss shares the property.

\begin{figure}[t]
\centering 
\includegraphics[width=0.345\textwidth]{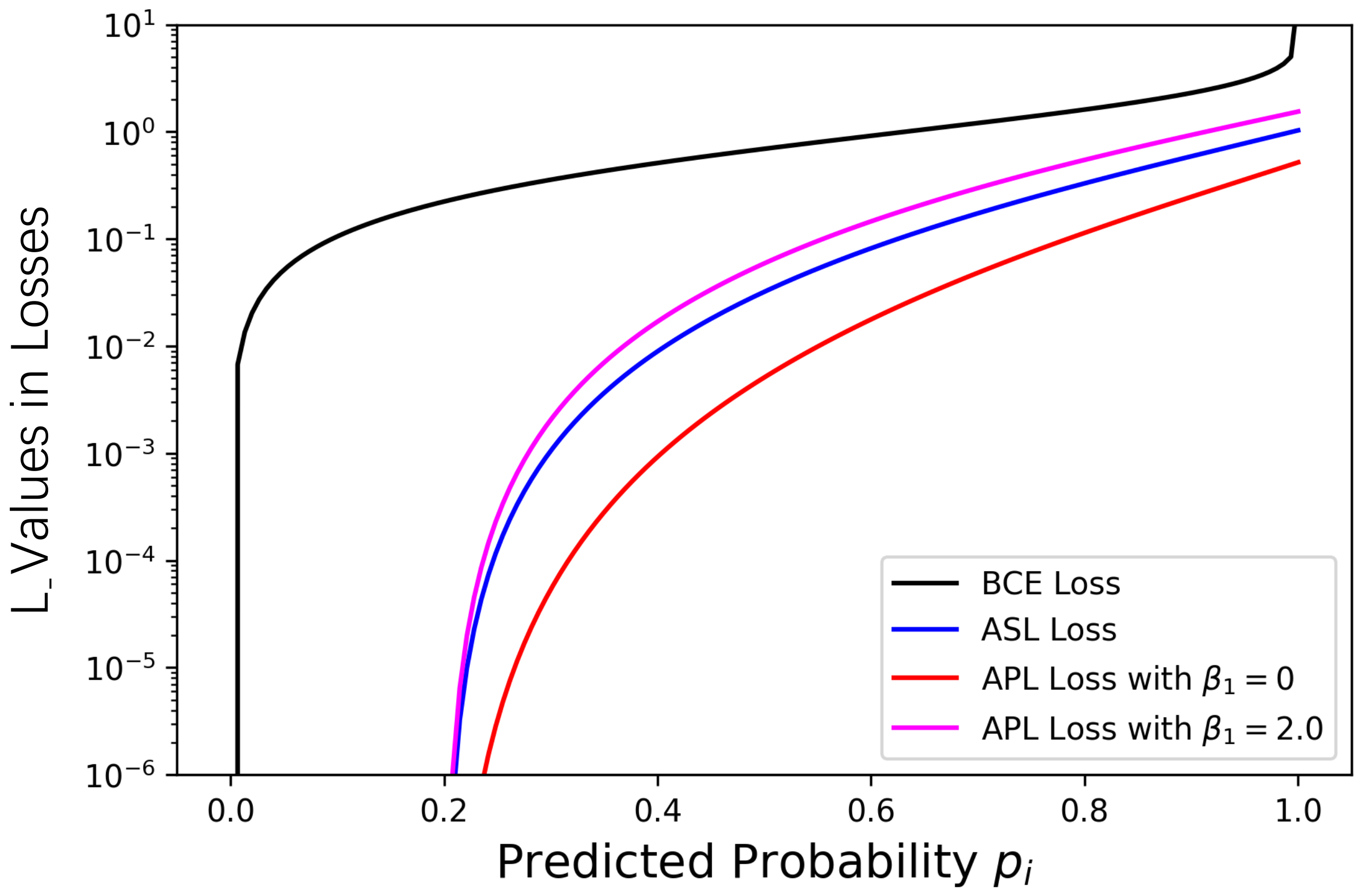}
\caption{Comparison of $L_-$ curves for different loss functions. $\gamma_-=2$ and $p_{th}=0.2$ for both ASL and APL for comparison.}
\label{pic_loss}
\end{figure}
By drawing $L_-$ loss curves in Figure \ref{pic_loss}, we can directly tell that easy negative classes whose prediction probabilities are lower than the threshold $p_{th}$ have no effect on training,
and adjusting the leading coefficient $\beta_1$ is beneficial to further optimize the loss ratio of positive and negative classes. 
Besides, adjusting the leading coefficient $\beta_1$ can theoretically alleviate the mislabelling problem, which will be further discussed in Section \ref{S3}. 

\subsection{Gradient Analysis}\label{S3}
We separately analyze gradients for positive and negative classes described in $L_{\rm APL}$ for clarity. 
For positive classes, without loss of generality, we set $\gamma^+=0$ and the gradient with respect to $p_i$ is
$
    -\frac{\partial L^+}{\partial p_i} = \alpha_1 + 2\alpha_2(1-p_i) + \sum_{j=2}^{\infty} (1-p_i)^j.
$
We observe that $\alpha_1$ provides a constant gradient regardless of $p_i$, and $\alpha_2$ maintains a linear correlation with $p_i$. But the rest terms are strongly suppressed as $p_i$ approaches $1$. Similar property holds for $\gamma^+>0$. Therefore, adjusting $\alpha_1$ and $\alpha_2$ is reasonable from the gradient perspective. 

\begin{figure}[t]
\centering 
\includegraphics[width=0.35\textwidth]{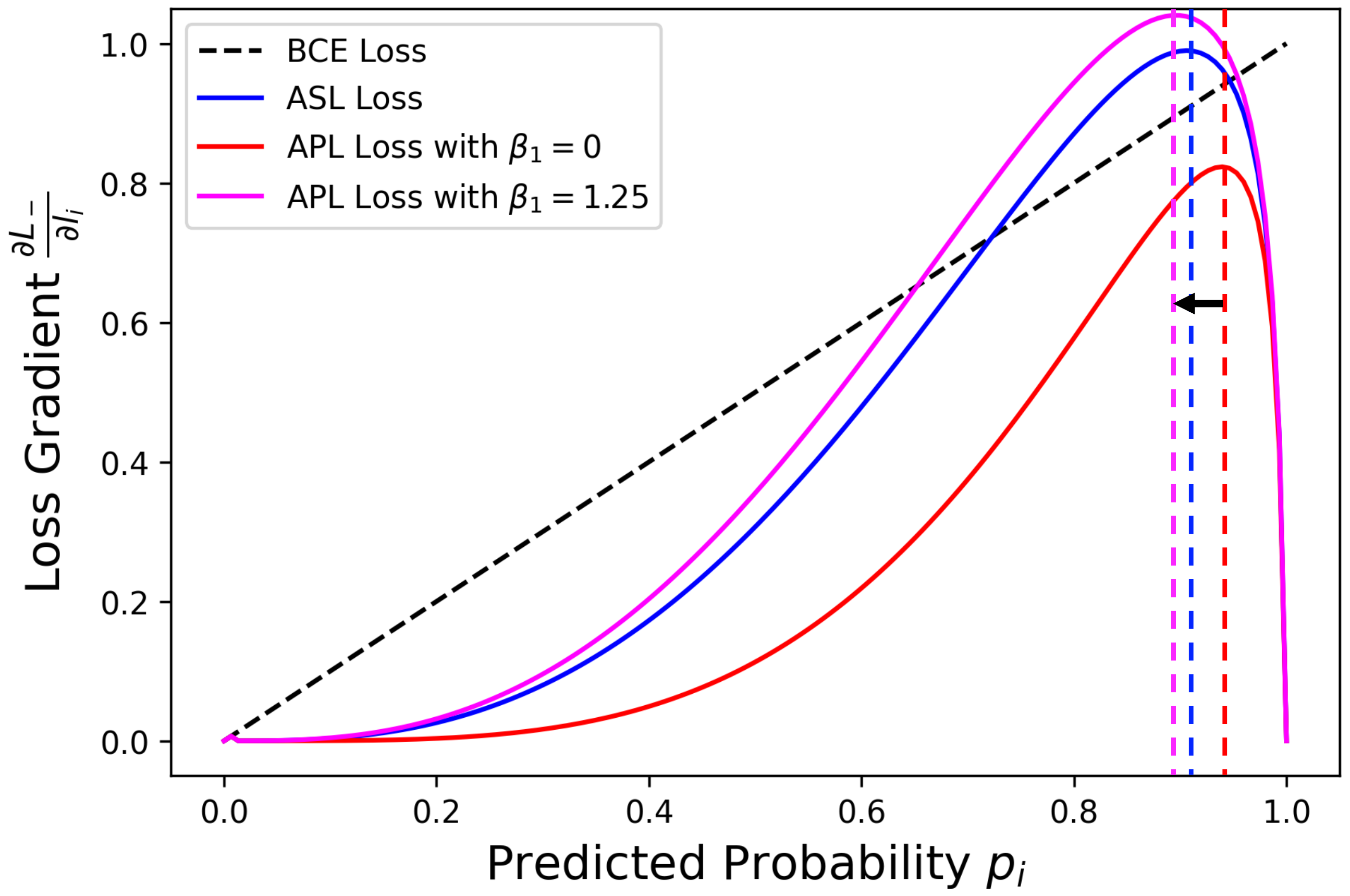}
\caption{We compare the gradients of $L_-$ losses of different loss functions. We set $\gamma_-=1.8$ and $p_{th}=0.01$ for both ASL and APL losses for comparison.}
\label{pic_gradient}
\end{figure}

For negative classes, the gradient with respect to logits $l_i$ is 
$
 \frac{\partial L^-}{\partial l_i} = \frac{\partial L^-}{\partial p_i} \frac{\partial p_i}{\partial l_i} = p_i(1-p_i)p_{res}^{\gamma^-}[ \frac{1}{1-p_{res}} - \frac{\gamma^- \log (1-p_{res})}{p_{res}} + (\beta_1-1)(\gamma^-+1) ].
$
We can find that adjusting $\beta_1$ can theoretically alleviate the mislabelling problem. Specifically, following previous works \cite{DBLP:conf/iccv/RidnikBZNFPZ21}, very hard negative samples are suspected to be mislabelled, i.e., predicted probability $p_i > p^*$ where $p^*$ is the solution to $\frac{\partial}{\partial p_i}(\frac{\partial L^-}{\partial l_i})=0$. However, simply changing $\gamma^-$ has little effect on shifting the solution $p^*$ when keeping $\beta_1$ as the default value $1$. But as shown in Figure \ref{pic_gradient}, slightly changing $\beta_1$ can greatly alter $p^*$ and the amplitude of the gradient curves. This finding might provide some insights to mitigate the mislabelling problem. Related experimental results on the AAPD \cite{DBLP:conf/coling/YangSLMWW18} dataset are reported in Section \ref{aapd2}.

\subsection{Parameter Interaction Analysis}\label{S4}
We validate that the polynomial coefficients (i.e., $\alpha_{1,2}$) can recalibrate the asymmetric focusing hyperparameters (i.e., $\gamma^+$).
For an imbalanced dataset, people usually keep the weight of rare positive classes unchanged and down-weight the negative classes, i.e., set $\gamma^+=0$ and fine-tune $\gamma^->0$. Assume that we have already found the best $\gamma^-$, but the target $\gamma^+=1$ is better. Then theoretically, we will find that $\alpha_1=0$ when we alter it, which means the first-order polynomial $\alpha_1(1-p)$ is an inappropriate term. 
By now, we can choose to either set $\gamma^+=1$ to raise the order for all polynomials or continue to alter $\alpha_2$ for better performance.
The difference between the two choices is
$
   \sum_{i=1}^{\infty} \frac{1}{i} (1-p)^{i+1} - \sum_{i=2}^{\infty} \frac{1}{i} (1-p)^i  = \frac{1}{2}(1-p)^2 +
  \frac{1}{6}(1-p)^3 + \frac{1}{12}(1-p)^4 + \cdots.
$
We can see that (i) the second choice to adjust $\alpha_2$ maintains smaller coefficients for polynomial bases; (ii) the difference between the same polynomial base decreases rapidly with the increase of polynomial order; (iii) high-order terms bring less gradients. 
Thus, \textbf{we need to continue to adjust $\alpha_2$ at most}, and the original smaller coefficient could reach a better solution. 
Besides, to obtain maximal gain with minimum parameter-tuning, we mildly suggest to compare the results of $\gamma^+=0$ and $\gamma^+=1$ after finding the optimal $\gamma^-$. Related experimental results on the MS-COCO dataset \cite{DBLP:conf/eccv/LinMBHPRDZ14} are reported in Section \ref{image}.

\section{Experiments}\label{res}
\subsection{Text Classification: AAPD Dataset}\label{aapd2}
\noindent\textbf{Dataset and Metrics.} We use the arXiv academic paper dataset AAPD \cite{DBLP:conf/coling/YangSLMWW18}, which contains 54 labels. Following the previous work \cite{DBLP:conf/emnlp/XiaoHCJ19}, we adopt both the precision at top \text{K} ($P@k$) and the normalized discounted cumulated gains at top \text{K} ($nDCG@K$) as our evaluation metrics. 

\noindent\textbf{Main Results.} 
Since our APL-based solution does not involve model architecture modifications, our APL can be combined with different structures to improve the performance. 
Here, we employ LSAN \cite{DBLP:conf/emnlp/XiaoHCJ19} model that performs best in the open-source codes as our baseline. 
For a fair comparison, to reduce the bias caused by devices,
we present both the original LSAN results and our re-implemented results. 
Our experiments are conducted on NVIDIA A100 GPU with CUDA version being 11.2 and PyTorch version being 1.9.0. 

We firstly re-implement the LSAN model with BCE loss and report the results in Table \ref{AAPD}. 
We further implement the ASL loss for comparison. In detail, we set $\gamma^+=0$, $p_{th}=0.05$ and tune $\gamma^-$ by grid search from $1$ to $5$. We obtain $85.31\%$, $85.73\%$, $86.15\%$, $85.83\%$ and $85.73\%$ P@1 metric scores respectively, and we therefore choose $\gamma^-=3$. By setting $\gamma^-=3$, we also change $\gamma^+$ from $1$ to $4$, and obtain $85.52\%, 85.42\%, 84.27\%$ and $83.65\%$ P@1 metric scores accordingly, which means $\gamma^+=0$ and $\gamma^-=3$ are the best choices for ASL loss. 
We consider this outcome as the ASL results reported in Table \ref{AAPD}.

\begin{table}
\centering
\resizebox{.80\columnwidth}{!}{
\begin{tabular}{lccccc}
\toprule
\textbf{Method}  & P@1 & P@3 & G@3 & G@5\\
\midrule
LSAN with BCE    &  85.28 & 61.12 & 80.84 & 84.78 \\
\midrule
LSAN*  & 85.10  & 60.80 & 80.55 & 84.52  \\
ASL Loss   & 86.15  & 61.01 & 80.90 & 84.50  \\
APL Loss   &  \textbf{86.56} & \textbf{61.35} & \textbf{81.34} & \textbf{84.86}  \\
\bottomrule
\end{tabular}}
\caption{Results (\%) on AAPD test set. G@K is abbreviated for $nDCG@K, K=3,5$. * denotes re-implemented results. \emph{Improvement} is compared with the results of LSAN*.}
\label{AAPD}
\end{table}

We then adjust the leading polynomial coefficient $\alpha_1$ of our APL with the above hyperparameters. 
In detail, the result when $\alpha_1=1$ is the benchmark since this is the result of the ASL loss. 
We increase the value of $\alpha_1$, we can observe that the metric values are rising consistently, and we report the results when $\alpha_1=2.5$ in Table \ref{AAPD}.
Results show that adjusting the coefficient of the first-order polynomial is effective. 

\begin{table}
\centering
\resizebox{.80\columnwidth}{!}{
\begin{tabular}{lcccc}
\toprule
\textbf{Model}  & \multicolumn{2}{c}{\textbf{Dev}} &   \multicolumn{2}{c}{\textbf{Test}} \\
\cmidrule(r){2-3} \cmidrule(r){4-5}
    & Ign $F_1$ & $F_1$ & Ign $F_1$ & $F_1$ \\
\midrule
MRN with ASL & 56.62 & 58.69 & 56.19 & 58.46 \\
\midrule
MRN with APL & \textbf{56.96} & \textbf{58.89} & \textbf{56.43} & \textbf{58.60} \\
\bottomrule
\end{tabular}}
\caption{Results ($\%$) on the dev and test sets of DocRED.}
\label{DocRED}
\end{table}

Another finding is that based on the ASL setting, i.e., $\alpha_1 = 1$ and we set $\beta_1 = 1.4$, we obtain the results $P@1=85.94\%$, $p@3=61.81\%$, $nDCG@3=81.62\%$, and $nDCG@5=85.03\%$. Compared with ASL results in Table \ref{AAPD}, we get $-0.21\%$ lower scores on precision at Top $1$ and achieve performance improvement of $+0.8\%, +0.72\%$ and $+0.53\%$ on the latter three metrics. 
The reason might be that the number of paper subjects could be limited when labeling, and some positive labels could be mislabelled as negative labels. Then, raising $\beta_1$ mitigates this issue as it would shift $p^*$ to the left as described in Section \ref{S3}. However, it would affect the optimization of some ordinary hard negative samples, resulting in the reduction of $P@1$. 

\subsection{Relation Extraction: DocRED Dataset}
\noindent\textbf{Dataset and Metrics.} We adopt the large-scale human-annotated document-level relation extraction dataset DocRED \cite{DBLP:conf/acl/YaoYLHLLLHZS19}, and we adopt the Ign F1 and F1 as the metrics.

\noindent\textbf{Main Results.}
We choose the recent open-sourced MRN model \cite{DBLP:conf/acl/LiXLFRJ21} as our baseline, which leverages the ASL loss and GloVe \cite{DBLP:conf/emnlp/PenningtonSM14} word embedding.
In detail, they adopt $\gamma^+=1, \gamma^-=4$ and $p_{th}=0.05$ for ASL loss. 
Following the same setting, we adjust $\alpha_1$ for our APL loss. 
Considering that $\alpha_1=1$ can be regarded as the ASL baseline, we set $\alpha_1=1.4$ and report the test set results as shown in Table \ref{DocRED}. Although the multi-label problem is not significant on the DocRED dataset (7\%), our APL loss still achieves better results on both development set and test set. 

We also conduct experiments using the pre-trained language model BERT \cite{DBLP:conf/naacl/DevlinCLT19}. Our APL loss achieves $59.78\%$ Ign F1 and $61.86\%$ F1 scores on the test set, which are higher than MRN+BERT with $59.52\%$ Ign F1 and $61.74\%$ F1 score. This shows that our APL loss can be effective using different word embeddings.

\subsection{Image Classification: MS-COCO and NUS-WIDE}\label{image}
\noindent\textbf{Dataset and Metrics.} We conduct experiments on MS-COCO \cite{DBLP:conf/eccv/LinMBHPRDZ14} and NUS-WIDE \cite{DBLP:conf/civr/ChuaTHLLZ09} multi-label image classification datasets.
Following the recent work ML\_Decoder \cite{DBLP:journals/corr/abs-2111-12933}, we present the mean average precision (mAP) score.

\noindent\textbf{Main Results.} We choose the ML\_Decoder model using ASL loss as the baseline and the results are reported in Table \ref{ImageResults}.
For the MS-COCO dataset that contains $122,218$ images with $80$ labels, we firstly report the original ASL \cite{DBLP:conf/iccv/RidnikBZNFPZ21} and ML\_Decoder results, where they both adopt $\gamma^+=0$, $\gamma^-=4$, $p_{th}=0.05$ ,and use TResNet-L \cite{DBLP:conf/wacv/RidnikLNBSF21} as the backbone. 

\begin{table}
\centering
\resizebox{.85\columnwidth}{!}{
\begin{tabular}{lc|c}
\toprule
\textbf{Method}  & MS-COCO & NUS-WIDE \\
\midrule
ASL Loss &  86.6 &  - \\
ML\_Decoder with ASL &  90.0 &  31.1 \\
\midrule
ML\_Decoder*  & 89.81  & 30.89   \\
APL Loss   &  \textbf{90.08} & \textbf{31.27}   \\
\bottomrule
\end{tabular}}
\caption{mAP scores on MS-COCO and NUS-WIDE test sets. * denotes re-implemented results.}
\label{ImageResults}
\end{table}

We firstly test different $\alpha_1$ for APL loss based on the above settings, and we notice that $\alpha_1=0$ achieves the best mAP scores, which indicates that $\gamma^+=0$ is not the best parameter for this model. As discussed in Section \ref{S4}, we conduct the following experiments:

\noindent (i) We further adjust $\alpha_2$ by fixing $\alpha_1=0$ and find that increasing $\alpha_2$ can further consistently enhance the performance. We report this result in Table \ref{ImageResults} by setting $\alpha_1=0$ and $\alpha_2=3.0$. 
Particularly, tuning $\alpha_2$ brings less improvement than tuning $\alpha_1$, which also shows that it is right to give priority to adjusting the leading polynomial coefficient. 

\noindent (ii) We only raise $\gamma^+=1$, and the performance is improved from $89.81\%$ to $89.99\%$. We further tune $\alpha_1 \in \{2.0, 2.5, 3.0, 3.5\}$ and obtain $90.00\%$, $90.02\%$, $90.02\%$, $90.02\%$ mAP scores. Raising $\alpha_1$ from $2.5$ to $3.5$ brings no additional improvement. 
We believe the reason might be that the coefficients of the same-order polynomial bases are large than those of the original Taylor expansion, resulting in a lower proportion of the gradient of the quadratic polynomial. Therefore, the effect of adjusting $\alpha_1$ is smaller when $\gamma^+>0$. 

For the NUS-WIDE dataset which is the multi-label zero-shot learning scenario  with $925$ seen labels from Flicker user tags and $81$ unseen human-annotated labels, 
the original ML\_Decoder uses the TResNet-L as the backbone and leverages the ASL loss with $\gamma^+=0, \gamma^-=4, p_{th}=0.05$ for optimization in Table \ref{ImageResults}.
We re-implement the ML\_Decoder model and we adjust $\alpha_1 \in \{ 1.2, 1.3, 1.4, 1.5 \}$ and obtain $31.10\%, 31.27\%, 31.12\%$ and $31.09\%$ mAP scores, respectively. We present the result using $\alpha_1=1.3$ in Table \ref{ImageResults}. Results show that our APL loss could further enhance the model performance on the multi-label zero-shot scenario. 

\section{Conclusion}
In this paper, we propose an asymmetric polynomial loss (APL) for multi-label classification.
Through conducting Taylor expansion on the BCE loss, our APL loss maintains high flexibility to tailor the polynomial coefficients for various tasks and models. 
Moreover, our APL loss enables us to alleviate the positive-negative imbalance problem by leveraging the asymmetric focusing mechanism. 
The polynomial coefficients and asymmetric focusing parameters are highly correlated both for positive and negative classes. 
Extensive experiments verity the effectiveness of our APL loss.

\section{Acknowledgement}
This work was sponsored by the National Natural Science Foundation of China (NSFC) grant (No.
62106143), and Shanghai Pujiang Program (No. 21PJ1405700).

\vfill\pagebreak

\bibliographystyle{IEEEbib}
\bibliography{strings}

\end{document}